%% file: root.tex
\documentclass[letterpaper, 10 pt, conference]{ieeeconf}  

\IEEEoverridecommandlockouts                              

\title{\LARGE \bf
SafeDMPs: Integrating Formal Safety with DMPs for Adaptive HRI 
}
\author{
Pranav Tiwari$^{*}$, Soumyodipta Nath$^{*}$, Ravi Prakash$^{\orcidlink{0000-0002-9058-434X}}$ \\
{\small \url{https://tiwari-pranav.github.io/SafeDMPs/}}%
\thanks{$^{*}$Denotes equal contribution.}%
\thanks{All authors are with the Cyber Physical Systems, Indian Institute of Science, Bengaluru, 560012, India. 
(e-mail: \{\texttt{pranavtiwari, soumyodiptan, ravipr}\}\texttt{@iisc.ac.in}).}
\thanks{Accepted to the IEEE International Conference on Robotics and Automation (ICRA), 2026.}
}

\usepackage{amssymb}
\usepackage{amsmath} 
\usepackage{graphicx} 
\usepackage{booktabs} 
\usepackage{subcaption}
\usepackage{algorithm}
\usepackage{algorithmic}
\usepackage{xcolor}
\usepackage{cite}
\usepackage[colorlinks=true, citecolor=cyan]{hyperref}
\usepackage{pifont} 
 \usepackage{multirow}
\usepackage{makecell}   
\newcommand{\cmark}{\ding{51}}%
\newcommand{\xmark}{\ding{55}}
\usepackage{stfloats}
\usepackage{orcidlink}

\begin{document}

\maketitle
\thispagestyle{empty}
\pagestyle{empty}

\begin{abstract}

Robots operating in human-centric environments must be both robust to disturbances and provably safe from collisions. Achieving these properties simultaneously and efficiently remains a central challenge. While Dynamic Movement Primitives (DMPs) offer inherent stability and generalization from single demonstrations, they lack formal safety guarantees. Conversely, formal methods like Control Barrier Functions (CBFs) provide provable safety but often rely on computationally expensive, real-time optimization, hindering their use in high-frequency control. This paper introduces SafeDMPs, a novel framework that resolves this trade-off. We integrate the closed-form efficiency and dynamic robustness of DMPs with a provably safe, non-optimization-based control law derived from Spatio-Temporal Tubes (STTs). This synergy allows us to generate motions that are not only robust to perturbations and adaptable to new goals, but also guaranteed to avoid static and dynamic obstacles. Our approach achieves a closed-form solution for a problem that traditionally requires online optimization.  Experimental results on a 7-DOF robot manipulator demonstrate that SafeDMPs is orders of magnitude faster and more accurate than optimization-based baselines, making it an ideal solution for real-time, safe, and collaborative robotics.

\end{abstract}

\section{INTRODUCTION}
\input{intro}

\section{Preliminaries}
\label{sec:preliminaries}
This section introduces the fundamental mathematical frameworks that underpin our approach to safe and adaptive human-robot interaction. 

\subsection{Dynamic Movement Primitives (DMPs) \cite{dmp}}
\label{subsec:dmps}

DMPs provide a unified framework for learning, representing, and generalizing robot motions from demonstrations. They model trajectories as a second-order dynamical system with a nonlinear forcing term that encodes the shape of the motion. The dynamics are governed by:
\begin{equation}
\tau^2 \ddot{x} = \alpha_z \big(\beta_z (g - x) - \tau\dot{x}\big) + f(z),
\label{eq:dmp_core}
\end{equation}
where $x(t) \in \mathbb{R}^d$ is the trajectory in $d$-dimensional space, $g \in \mathbb{R}^d$ is the goal state, $\tau>0$ is a temporal scaling factor controlling duration, $\alpha_z,\beta_z >0$ are gains (with $\beta_z=\alpha_z/4$ for critical damping), and $f(z)$ is the nonlinear forcing function that shapes the trajectory governed by a canonical monotonically decreasing phase variable
\begin{equation}
\tau \dot{z} = -\alpha_z z, \quad z(0)=1,
\label{eq:dmp_phase}
\end{equation}
From a demonstration, the target forcing function is obtained by rearranging Eq.~\eqref{eq:dmp_core} and is parameterized using a weighted combination of Gaussian basis functions.

This formulation enables flexible adaptation through multiple scaling mechanisms:

\paragraph{Stable Convergence} Attractor dynamics guarantees goal reaching even under perturbations. 

\paragraph{Spatial Scaling} 
Trajectories can be spatially transformed by modifying the initial position $y_0$ and goal $g$ where the forcing function automatically scales the trajectory shape to the new spatial extent.

\paragraph{Temporal Scaling}
The duration and speed of the motion are controlled by the scaling parameter $\tau > 0$, where $\tau > 1$ slows execution and $0 < \tau < 1$ speeds it up.
\label{eq:temporal_scaling}

\paragraph{Obstacle Avoidance}
DMPs can be extended with coupling terms for real-time obstacle avoidance:
\begin{equation}
\tau^2 \ddot{x} = \alpha_z (\beta_z (g - x) - \tau \dot{x}) + f(z) + f_{\text{coupling}}(x, \dot{x}, \text{obstacles}),
\label{eq:dmp_coupling}
\end{equation}
where $f_{\text{coupling}}$ generates repulsive forces from detected obstacles.

\subsection{Spatio-Temporal Tubes (STT) \cite{stt}}
\label{subsec:st_tubes}

STTs enforce state-dependent safety by constraining each state dimension $x_i(t)$ within time-varying bounds:
\begin{equation}
\rho_{i,L}(t) < x_i(t) < \rho_{i,U}(t), \quad \forall t \geq 0,
\end{equation}
without requiring online optimization by using a normalized error representation. Define the sum and difference of bounds:
\vspace{-0.5em}
\begin{align}
\rho_{i,s}(t) &= \rho_{i,U}(t) + \rho_{i,L}(t), \label{eq:bounds_sum} \\
\rho_{i,d}(t) &= \rho_{i,U}(t) - \rho_{i,L}(t), \label{eq:bounds_diff}
\end{align}

The normalized error for dimension $i$ is:
\begin{equation}
e_i(x_i,t) = 2 \rho_{i,d}^{-1}(t) \left( x_i(t) - \frac{1}{2} \rho_{i,s}(t) \right),
\label{eq:normalized_error}
\end{equation}
which maps the tube interior to $(-1, 1)$ and centers the error at zero when $x_i(t)$ is at the tube center.

The control function that enforces tube constraints is:
\begin{equation}
\xi_i(x_i,t) = 4 \rho_{i,d}^{-1}(t) \left( 1 - e_i(x_i,t)^2 \right),
\label{eq:control_function}
\end{equation}
which provides maximum control authority at the tube boundaries and minimum at the center.

The STT control law ensures finite-time convergence to the target set while maintaining tube constraints:
\begin{align}
u_i(x_i,t) &= -k \, \xi_i(x_i,t) \, \epsilon_i(x_i,t), \quad k > 0, 
\label{eq:stt_control_law} \\
\epsilon_i(x_i,t) &= \ln\left( \frac{1 + e_i(x_i,t)}{1 - e_i(x_i,t)} \right),
\label{eq:error_term}
\end{align}

This logarithmic error term $\epsilon_i$ ensures:
\begin{itemize}
    \item $\epsilon_i \to -\infty$ as $e_i \to -1$ (approaching lower bound)
    \item $\epsilon_i = 0$ when $e_i = 0$ (at tube center) 
    \item $\epsilon_i \to +\infty$ as $e_i \to +1$ (approaching upper bound)
\end{itemize}

This property makes STTs particularly effective for human-robot interaction, where responsiveness is critical and online optimization may be infeasible.

\section{Methodology}
\label{sec:methodology}

We propose a two-stage framework for generating robust, adaptive, and safe robot motion in dynamic environments. The framework consists of (i) learning nominal motion plans from demonstrations and (ii) enforcing robustness and safety during execution. Our method operates in the end-effector space, directly encoding cartesian trajectories to improve generalization and maintain explicit safety margins.

\subsection{DMPs as Nominal Motion Plans}
\label{subsec:nominal_learning}
Learning from demonstration (LfD) allows robots to acquire complex motor behaviors from kinesthetic teaching. We adopt Dynamic Movement Primitive (DMPs) \cite{dmp} as our nominal representation because they provide stable convergence guarantees, temporal flexibility (through the parameter $\tau$), and spatial generalization (see Sec.~\ref{subsec:dmps}). In addition, we can easily embed an internal feedback term that enables smooth recovery from deviations without requiring an external correction layer. This adaptability is particularly valuable when integrated with Spatio-Temporal Tubes (STTs), because the deviations caused by safety constraints can be absorbed into the primitive’s dynamics.

\begin{figure*}[!t]
\centering
    \includegraphics[width=\linewidth, height=0.35\linewidth]{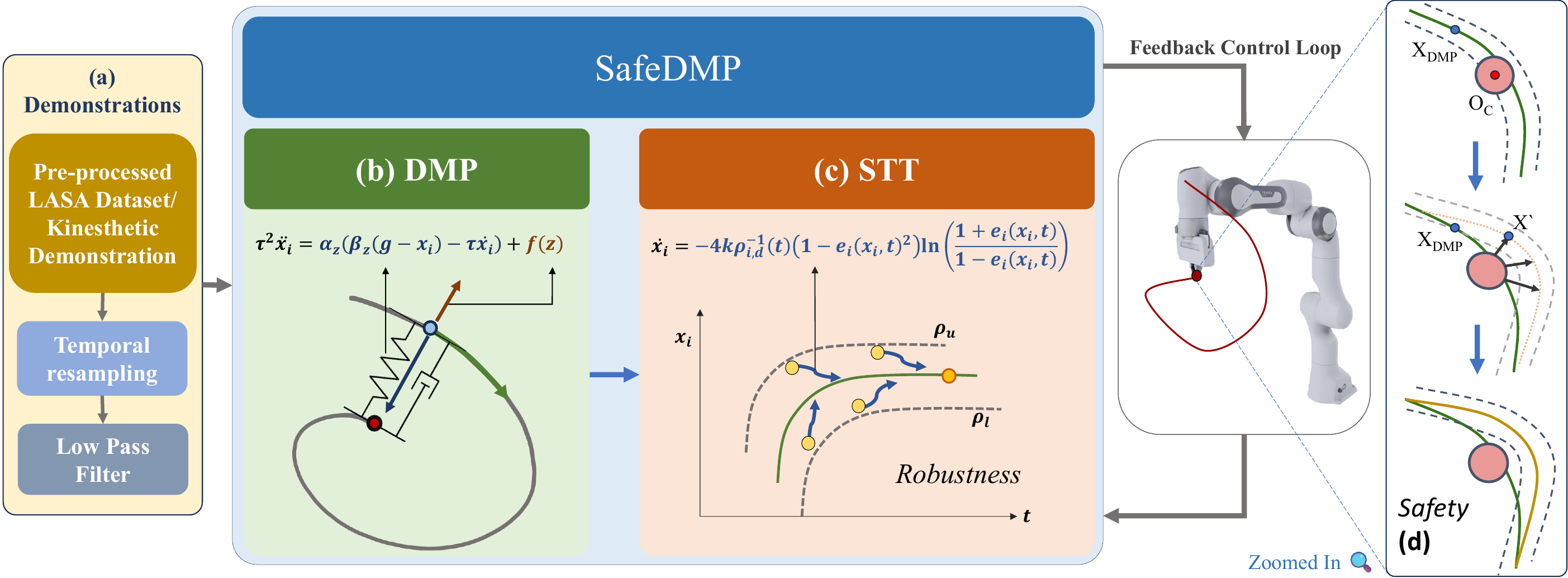}
    \caption{Proposed framework pipeline. (a) Demonstrations are recorded in end-effector space. (b) DMPs encode the nominal motion plan. (c) Spatio-Temporal Tubes (STTs) define a safe envelope around the DMP. (d) During execution, trajectory deformation is triggered when obstacles are encountered, and the system safely reroutes motion within the STT while ensuring convergence.}
    \label{fig:fig_proposed_framework}

    \vspace{-5mm}
\end{figure*}

\subsection{SafeDMPs: Enforcing Robustness and Safety via Spatio-Temporal Tubes}
\label{subsec:stt_control}

While DMPs ensure convergence and adaptability, they do not guarantee safety in dynamic environments. To address this, we integrate STTs \cite{stt} into the DMP execution layer. STTs define time-varying safety envelopes for each state dimension around the nominal trajectory (see Sec.~\ref{subsec:st_tubes}, Eq.~\eqref{eq:normalized_error}--\eqref{eq:error_term}), and enforce constraint satisfaction through a closed-form control law (Eq.~\eqref{eq:stt_control_law}) without requiring online optimization. Figure~\ref{fig:fig_proposed_framework} illustrates this integration.

The closed-loop system additively integrates the expressiveness of DMPs with the safety guarantees of STTs to yield a closed-form, optimization-free dynamical system (Fig.~\ref{fig:fig_proposed_framework}(d)). This structure adaptively regulates motion while preserving the distinct advantages of both frameworks. The DMP provides a critically damped, second-order attractor for robust goal convergence, while the STT introduces an obstacle-aware modulation term, $f_{\text{STT}}$, to handle environmental perturbations:
\begin{equation}
\tau^2 \ddot{x} = \alpha_z \big(\beta_z (g - x) - \tau\dot{x}\big) + f(z) + f_{\text{STT}}(z, x, \mathcal{O}).
\label{eq:dmp_stt_combined}
\end{equation}
The STT ensures forward invariance within a safe envelope ($\rho_{i,L}(t) < x_i(t) < \rho_{i,U}(t)$) using a diverging logarithmic error term that dominates near boundaries to strictly prevent constraint violations. When dynamic obstacles $\mathcal{O}$ arise, the STT safely deforms the nominal DMP trajectory ($\mathbf{x}_{\text{DMP}}$) based on the Euclidean distance $d$ to the obstacle center $\mathbf{o}_c$:
\begin{equation}
\mathbf{x}^{\prime} =
\begin{cases}
\mathbf{x}_{\text{DMP}}, & \text{if } d(\mathbf{x}_{\text{DMP}}, \mathbf{o}_c) > r_{\text{safe}}, \\
\mathbf{o}_c + \frac{\mathbf{x}_{\text{DMP}} - \mathbf{o}_c}{d(\mathbf{x}_{\text{DMP}}, \mathbf{o}_c)} \cdot r_{\text{clearance}}, & \text{otherwise}.
\end{cases}
\end{equation}
This rerouting strategy enforces a minimum clearance ($r_{\text{clearance}}$) upon breaching a buffer zone ($r_{\text{safe}}$), securely guiding the end-effector while the underlying DMP ensures goal convergence. While formulated for spherical constraints, replacing the point-to-center distance $d(\cdot)$ with a \textit{signed distance function} naturally extends this approach to arbitrary obstacle geometries.

\subsection{Adaptive Timing for Recovery}
To handle large deviations from the nominal path, we further adapt the temporal scaling parameter $\tau$:  
\begin{align}
\dot{\mathbf{e}} &= \alpha_e (\mathbf{x}_{\text{actual}} - \mathbf{x}_{\text{DMP}} - \mathbf{e}), \\
\tau &= \tau_{\text{nominal}} + k_c \|\mathbf{e}\|^2.
\end{align}
If deviations grow large, $\tau$ increases, slowing down execution and allowing more time for recovery; when deviations are small, $\tau$ decreases, speeding up execution. This ensures recovery remains smooth and bounded rather than abrupt (Fig.~\ref{fig:fig_proposed_framework}(d)).

\section{Experimental Evaluation}
\label{sec:experiments}

\subsubsection{Robot and Environment}
\begin{figure}[htpb]
    \centering
    \includegraphics[width=0.85\linewidth]{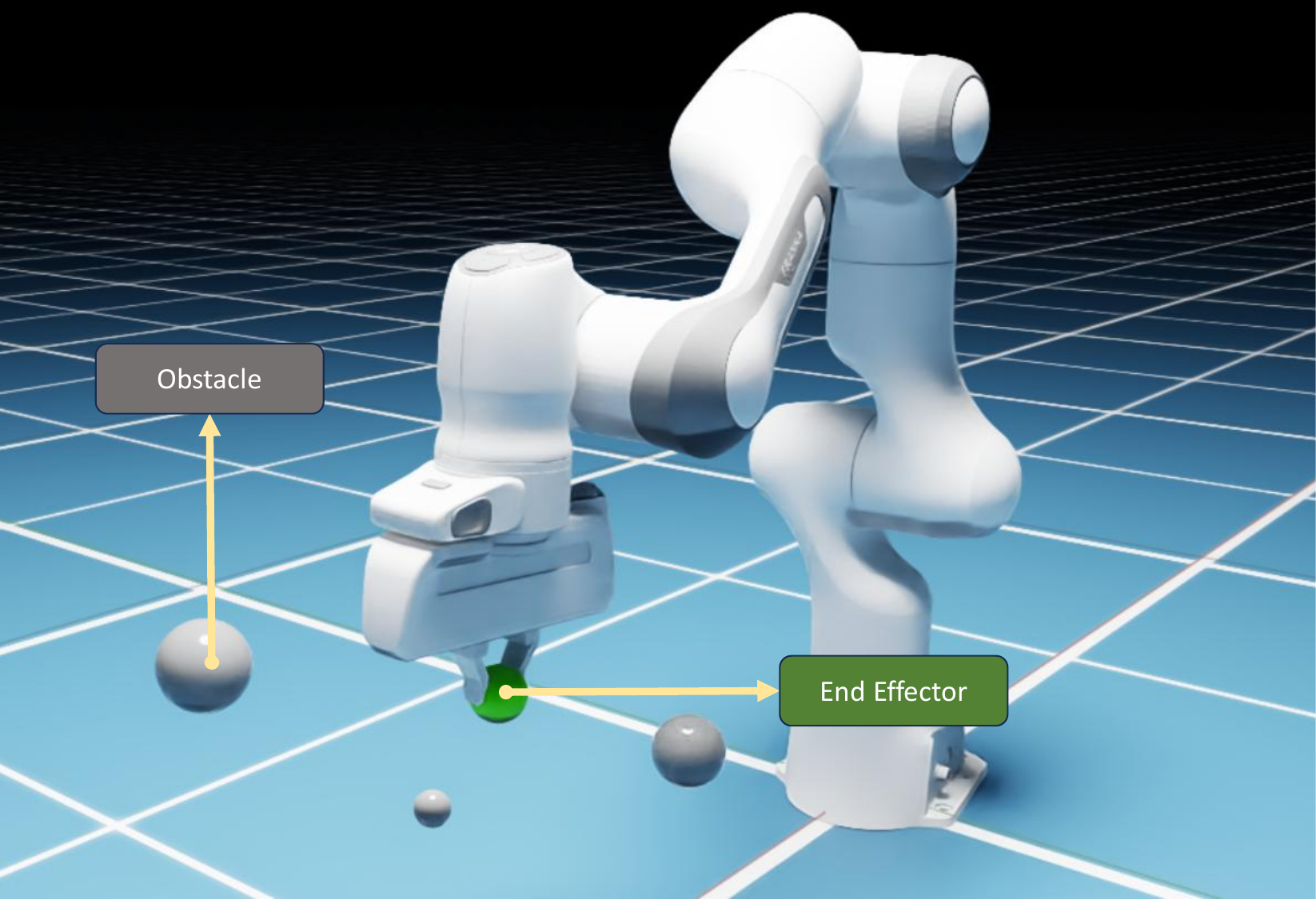}
    \caption{Simulation setup featuring the Franka robot performing tasks within a workspace containing obstacles.}
    \label{fig:franka_sim}

    \vspace{-5mm}
\end{figure}

All simulations were conducted using a model of the 7-DOF Franka Emika Panda  manipulator in Isaac Sim within a cubic workspace measuring $1.1 \times 1.1 \times 1.1$~m (Fig.~\ref{fig:franka_sim}). 
The controller ran at a frequency of 200~Hz ($\Delta t = 0.005$~s) 


\begin{algorithm}[H]
\caption{Integrated DMP-STT Motion Planning Framework with Embedded STT Safety Control}
\label{alg:dmp_stt}
\begin{algorithmic}[1]
\REQUIRE Reference trajectory $\mathbf{X}_{ref} \in \mathbb{R}^{N \times d}$, DMP parameters $\alpha$, $n_{features}$, STT parameters $K$, $\delta_\gamma$, obstacle $(\mathbf{o}_c, r_o)$, time step $\Delta t$
\ENSURE Executed safe trajectory 
\vspace{0.5em}
\STATE \textbf{Phase 1: DMP Learning}
\STATE Initialize: $\beta \gets \alpha/4$, $\alpha_z \gets \alpha/6$, $\alpha_e \gets \alpha/10$, $k_c \gets 2\alpha$ 
\STATE $\mathbf{x}_0 \gets \mathbf{X}_{ref}[0]$, \; $\mathbf{g} \gets \mathbf{X}_{ref}[-1]$
\STATE Learn Weights: $\mathbf{w} \gets {DMP}_{\text{train}}(X_{ref}, \beta, \alpha, \alpha_z)$

\vspace{0.5em}
\STATE \textbf{Phase 2: Real-time Execution}
\STATE Initialize: $z \gets 1$, $\mathbf{X} \gets \mathbf{x}_0$, $\dot{\mathbf{X}} \gets \mathbf{0}$, $\mathbf{e} \gets \mathbf{0}$

\WHILE{$\|\mathbf{X} - \mathbf{g}\| > \epsilon_{thresh}$}
    \vspace{0.2em}
    \STATE \textbf{DMP Step:}
    \vspace{0.2em}
    \STATE $\boldsymbol{\psi} \gets \exp(-\mathbf{h}_z \cdot (z - \mathbf{c}_z)^2)$
    \STATE $\mathbf{f}_{ext} \gets (\mathbf{g} - \mathbf{x}_0) \cdot z \cdot ((\boldsymbol{\psi}/\sum\boldsymbol{\psi})^T \mathbf{w})$
    \STATE $\ddot{\mathbf{X}} \gets \tfrac{1}{\tau^2}\big(\alpha(\beta(\mathbf{g} - \mathbf{X}) - \tau\dot{\mathbf{X}}) + \mathbf{f}_{ext}\big)$
    \vspace{0.2em}
    \STATE \textbf{Embedded STT Safety Control:}
    \vspace{0.2em}
    \STATE $\mathbf{X}_{target} \gets \mathbf{X} + \dot{\mathbf{X}} \cdot \Delta t$
    \STATE $\mathbf{x}_{safe} \gets \mathbf{X}_{target}$

    \IF{obstacle present}
        \STATE $\mathbf{v}_{ot} \gets \mathbf{X}_{target} - \mathbf{o}_c$, \quad $d_{ot} \gets \|\mathbf{v}_{ot}\|$
        \STATE $clearance \gets r_o + 0.5\delta_\gamma$
        \IF{$d_{ot} \leq clearance$}
            \STATE $\mathbf{x}_{safe} \gets \mathbf{o}_c + \frac{\mathbf{v}_{ot}}{d_{ot}} \cdot clearance$
        \ENDIF
    \ENDIF

    \STATE Error: $\boldsymbol{\epsilon} \gets \frac{\mathbf{X}_{current} - \mathbf{x}_{safe}}{0.5 \delta_\gamma}$
    \STATE Clip: $\boldsymbol{\epsilon} \gets \text{clip}(\boldsymbol{\epsilon}, -0.99, 0.99)$
    \STATE Transform: $\boldsymbol{\epsilon}_t \gets \log\!\left(\frac{1+\boldsymbol{\epsilon}}{1-\boldsymbol{\epsilon}}\right)$
    \STATE Control: $\mathbf{u}_{stt} \gets -K \cdot \tfrac{4}{\delta_\gamma (1 - \boldsymbol{\epsilon}^2)} \cdot \boldsymbol{\epsilon}_t$
    \STATE $\mathbf{X}_{desired} \gets \mathbf{X}_{target} + \mathbf{u}_{stt} \cdot \Delta t$

    \STATE \textbf{Robot Control:}
    \STATE Move to $\mathbf{X}_{desired}$
    \STATE $\mathbf{X}_{current} \gets \text{get\_current\_ee\_position}()$

    \STATE \textbf{Update DMP State:}
    \STATE $\mathbf{e} \gets \mathbf{e} + \alpha_e (\mathbf{X}_{current} - \mathbf{X}) \Delta t$
    \STATE $\tau \gets \tau_{base} + k_c \|\mathbf{e}\|^2$
    \STATE $z \gets z - \frac{\alpha_z z \Delta t}{\tau}$
    \STATE $\mathbf{X} \gets \mathbf{X} + \dot{\mathbf{X}}\Delta t + 0.5 \ddot{\mathbf{X}} \Delta t^2$
    \STATE $\dot{\mathbf{X}} \gets \dot{\mathbf{X}} + \ddot{\mathbf{X}} \Delta t$
\ENDWHILE

\end{algorithmic}
\end{algorithm}

\vspace{-5mm}

\subsubsection{Motion Data and Nominal Plan}
We utilized the LASA handwriting dataset \cite{khansari2011learning} as the source of example motions. The original 2D demonstrations were lifted into 3D space (with a fixed $z$-height), randomly rotated, temporally resampled, and smoothed with a low-pass filter. These processed trajectories were encoded as DMPs using 25 basis functions to serve as the nominal, obstacle-free motion plan for all subsequent simulation experiments. 

All experiments were conducted on an Intel Core i7-12700K workstation with 16 GB RAM. The complete set of hyperparameters and implementation details for both hardware \& software are provided in the repository at \href{https://github.com/Tiwari-Pranav/ghost-In-the-Arm/tree/main}{\textit{GitHub}}. All experimental videos are available on the \href{https://tiwari-pranav.github.io/SafeDMPs/}{\textit{project website}}.



\subsection{Baselines for Comparison}
\label{subsec:baselines}

We benchmark the performance of our proposed SafeDMPs against two widely used methods:  
\begin{itemize}
    \item \textbf{NODE--CLF--CBF} \cite{nawaz2024reactive}: learning-based dynamics combined with control barrier functions, offering formal safety guarantees via online optimization.  
    \item \textbf{DMP--APF} \cite{dmp_based_oa}: nominal motion encoding with artificial potential fields for reactive avoidance.  
\end{itemize}

\begin{table*}[htpb]
\centering
\caption{\large Qualitative comparison and Quantitative evaluation of motion generation methods.}
\label{tab:results}
\renewcommand{\arraystretch}{1.2}
\setlength{\tabcolsep}{5pt}
\begin{tabular}{lcccccccccc}
\toprule
\multirow{2}{*}{\textbf{Method}} & 
\multicolumn{4}{c}{\textbf{Qualitative Properties}} & 
\multicolumn{6}{c}{\textbf{Quantitative Metrics}} \\
\cmidrule(lr){2-5} \cmidrule(lr){6-11}
 & \makecell{\textbf{Formal}\\\textbf{Safety}} & 
   \makecell{\textbf{Robust}\\\textbf{Recovery}} & 
   \makecell{\textbf{No Oscill.}\\\textbf{/ Minima}} & 
   \makecell{\textbf{Real-Time}\\\textbf{Efficiency}} & 
   \makecell{\textbf{Exec.}\\\textbf{Time (s)}} & 
   \makecell{\textbf{Mem.}\\\textbf{(MB)}} & 
   \makecell{\textbf{MAE}\\\textbf{Nom.}} & 
   \makecell{\textbf{MAE}\\\textbf{Pert.}} &
   \makecell{\textbf{Conv.}\\\textbf{T OA}} & 
   \makecell{\textbf{Conv.}\\\textbf{T Pert.}} \\
\midrule
NODE--CLF--CBF \cite{nawaz2024reactive}          & \cmark & \xmark & \cmark & \xmark & 0.3207   & 600    & 0.2244 & 0.3921 & 0.411 & 1.294 \\
NODE--CLF--CBF (JIT) \cite{nawaz2024reactive}    & \cmark & \xmark & \cmark & \xmark & 1.8e-3   & 0.311 & 0.2244 & 0.3921 & 0.411 & 1.294 \\
DMP--APF \cite{dmp_based_oa}                & \xmark & \xmark & \xmark & \cmark & 6.7e-4       & 0.0208    & 0.02304  & 0.0228   & $\-$ & 0.411 \\
DMP--APF (JIT) \cite{dmp_based_oa}          & \xmark & \xmark & \xmark & \cmark & 2.3e-4       & 0.0208    & 0.02304    & 0.0228   & $\-$ & 0.411 \\
\textbf{SafeDMPs}                             & \cmark & \cmark & \cmark & \cmark & \textbf{1.03e-4} & \textbf{0.109} & \textbf{0.0113} & \textbf{0.0531} & \textbf{0.6325} & \textbf{0.025} \\
\textbf{SafeDMPs (JIT)}                       & \cmark & \cmark & \cmark & \cmark & \textbf{3.7e-5} & \textbf{0.161} & \textbf{0.0113} & \textbf{0.0531} & \textbf{0.6325} & \textbf{0.025} \\
\bottomrule
\end{tabular}

\begin{flushleft}
{\footnotesize \textit{\textbf{Qualitative metrics:}} $\checkmark$ Formal Safety: proven forward invariance; $\checkmark$ Robust Recovery: bounded error convergence post-perturbation; $\checkmark$ No Oscill./Minima: zero instances across 100 trials. 
\textit{\textbf{JIT implementation:}} Uses \texttt{numba} (SafeDMP, DMP--APF, CLF-CBF) and PyTorch JIT (NODE), excluding initial compilation overhead to capture steady-state control.}
\end{flushleft}

\vspace{-2mm}
\end{table*}

\begin{figure*}[!ht]
    \centering
    \includegraphics[width=0.9\textwidth, height=0.6\linewidth]{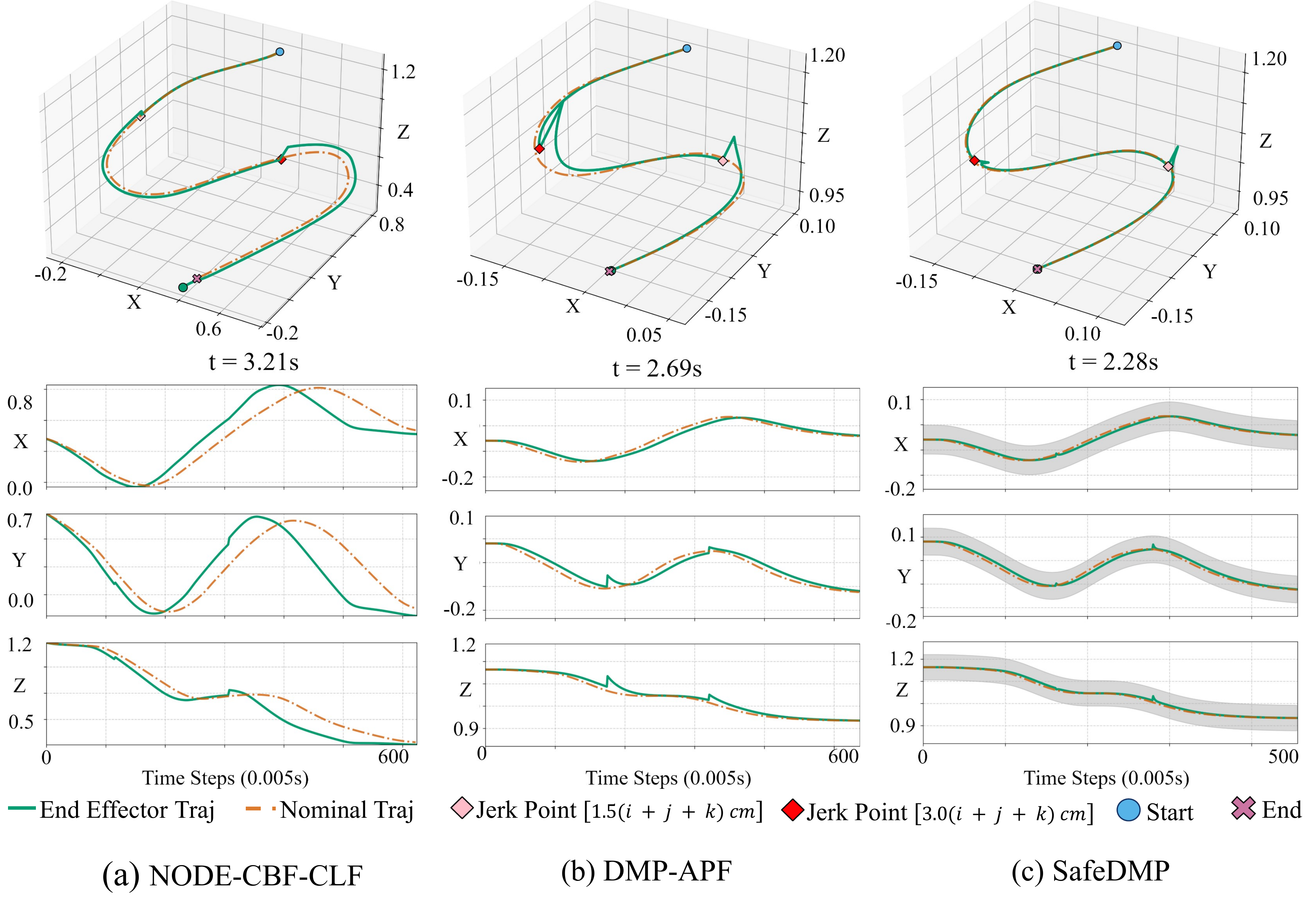}
    \caption{Perturbation response across methods. 
    The plots illustrate the evolution of the end-effector trajectory over time (top) and the corresponding $X$, $Y$, and $Z$ coordinates (bottom) under external disturbances.}
    \label{fig:perturb}

    \vspace{-2.5mm}
\end{figure*}

\subsection{Experimental Tasks and Evaluation Metrics}
\label{subsec:tasks}
To comprehensively evaluate each algorithm, we designed the following experimental tasks, complemented by quantitative measures that capture both efficiency and reliability.  

\subsubsection{Nominal Trajectory Reproduction}
Execution of the learned motion in an obstacle-free environment establishes a baseline for performance. The evaluation includes: i) \textbf{Execution Time (s):} The average time required for computing control commands per control loop to reflect computational efficiency. ii) \textbf{Memory Usage (MB):} The average memory footprint measured during execution. iii) \textbf{Mean Absolute Error | Nominal (m):} The deviation between the executed trajectory and the reference demonstration, quantifying imitation fidelity under nominal conditions.  

\subsubsection{Robustness}
Two sharp impulse perturbations are applied to the end-effector at identical instants across all algorithms. Evaluation focuses on: i) \textbf{Mean Absolute Error | Perturbation (m):} The deviation from the reference trajectory under disturbances, capturing resilience to perturbations. ii) \textbf{Convergence Time |  Perturbation (s)}: The average elapsed time for the perturbed trajectory to re-converge with the nominal path.

\subsubsection{Safety}
Safety is assessed in the presence of environmental constraints: i) \textbf{Static Obstacle Avoidance:} Obstacles are placed at random positions intersecting the nominal trajectory. The evaluation measures whether collisions are avoided and the smoothness of trajectory deformation. 
    ii) \textbf{Dynamic Obstacle Avoidance:} Obstacle moves with constant velocity on a path intersecting the nominal trajectory. Algorithms are evaluated on their ability to predict obstacle motion and generate proactive collision-free trajectories.  
    iii) \textbf{Convergence Time | Obstacle Avoidance (s):} The mean additional time required for the trajectory to return to the goal after executing obstacle avoidance maneuvers per obstacle.  

\begin{figure*}[htbp]
    \centering
    \includegraphics[width=0.9\textwidth, height=0.6\linewidth]{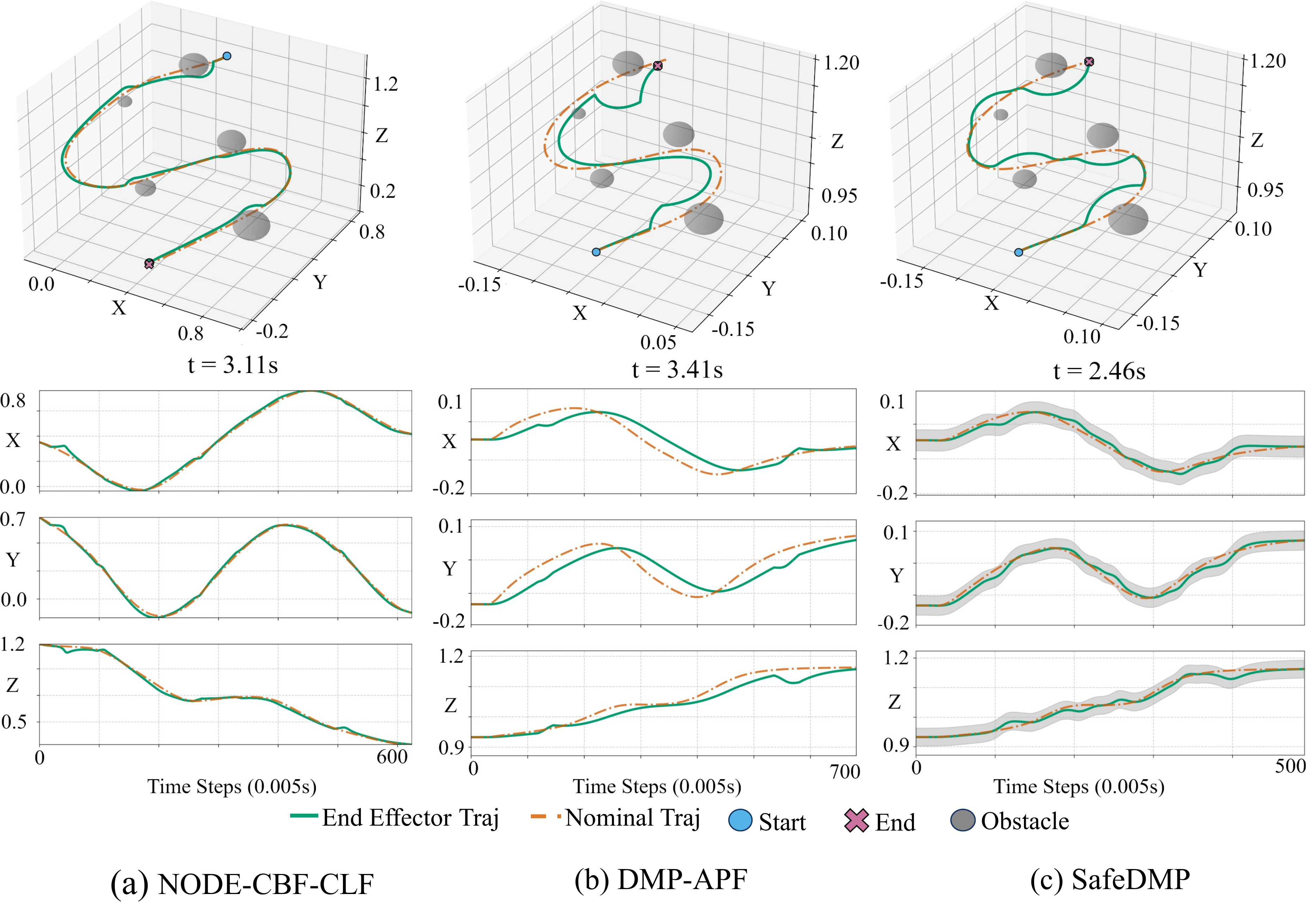}
    \caption{Obstacle avoidance across methods with static obstacles. 
    The plots show the evolution of the end-effector trajectory when static obstacles are introduced during execution (top) and the corresponding $X$, $Y$, and $Z$ coordinates (bottom). }
    \label{fig:obstacle_comparison}

    \vspace{-2.5mm}
\end{figure*}

\begin{figure*}[htbp]
    \centering
    \includegraphics[width=0.9\textwidth]{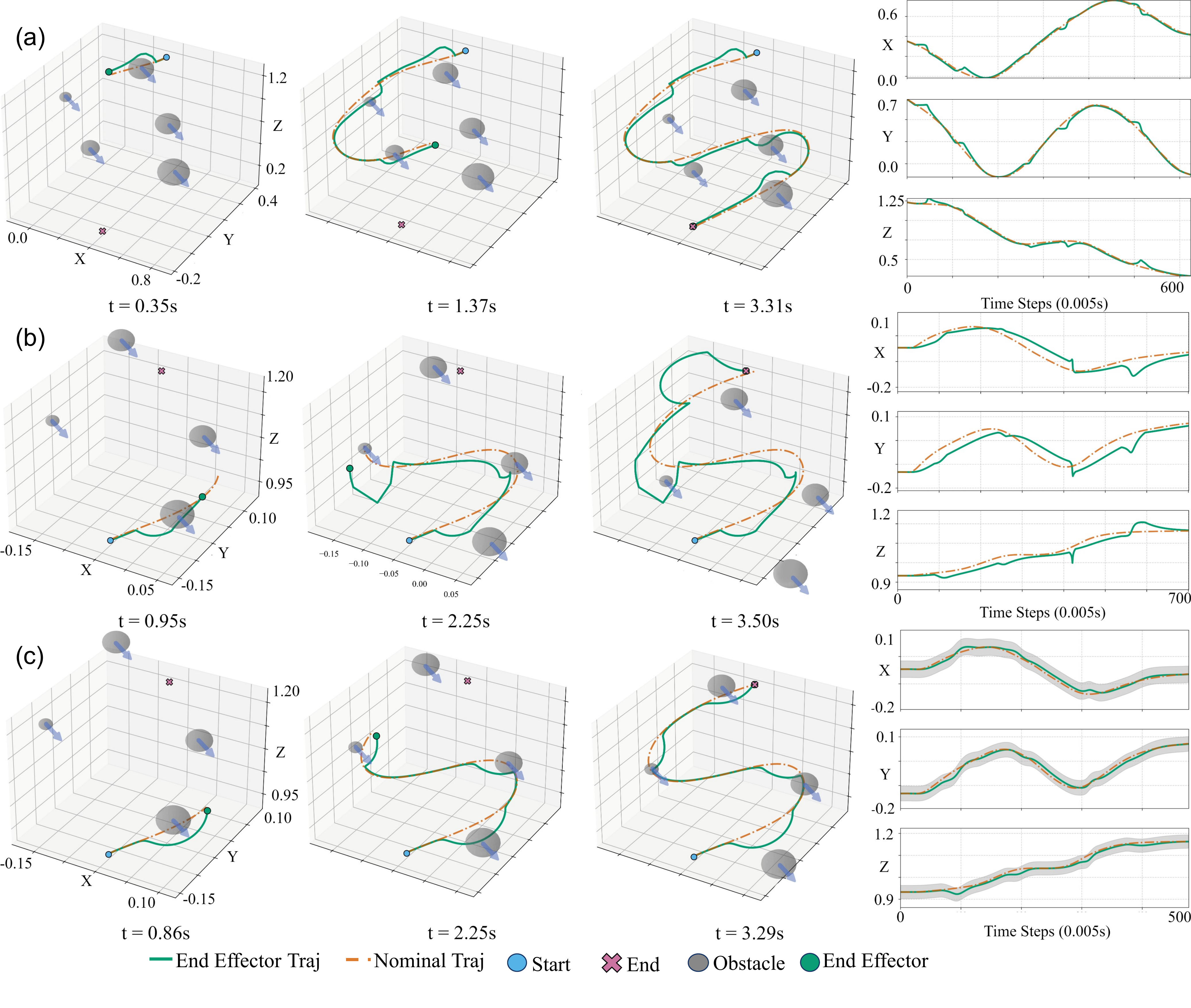}
    \caption{Obstacle avoidance across methods with moving obstacles. 
    The plots show three 3D frames (left) illustrating the evolution of the end-effector trajectory and the corresponding $X$, $Y$, and $Z$ coordinates (right).}
    \label{fig:moving_obstacle_comparison}
    
    \vspace{-2.5mm}
\end{figure*}

For clarity in visualization, the results presented in Figures~\ref{fig:perturb}--\ref{fig:moving_obstacle_comparison} depict the Cartesian path of the end-effector highlighted in Figure~\ref{fig:franka_sim}.

\begin{figure*}[htbp]
    \centering
    \includegraphics[width=0.9\textwidth, height=0.6\textwidth]{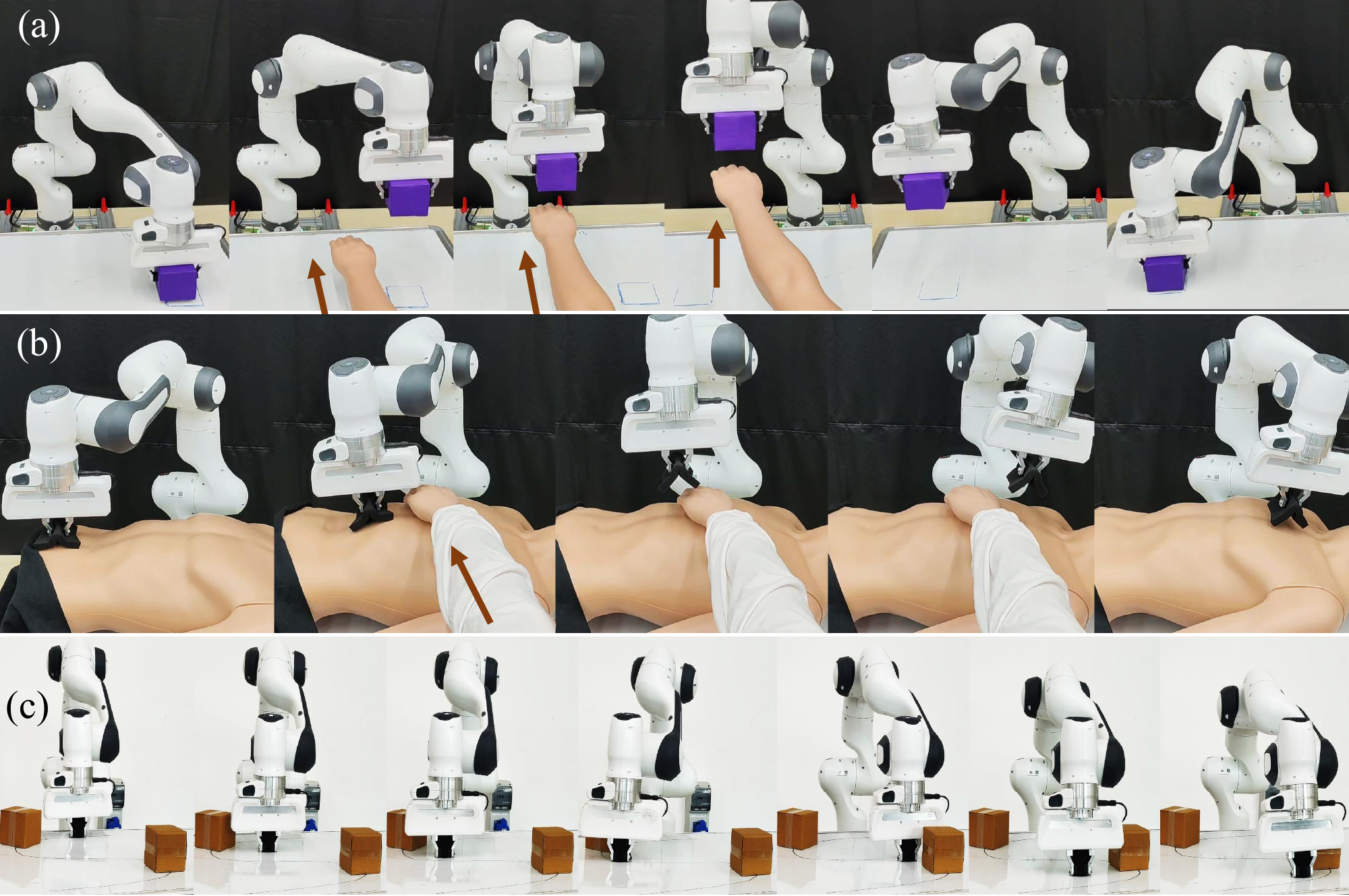}
    \caption{Hardware validation of the SafeDMP framework in human-robot interaction (HRI) scenarios. 
    \textbf{(a)} Pick-and-Place: The robot reactively deviates from its nominal trajectory to avoid a dynamically moving human hand.
    \textbf{ (b)} Assistive Wiping: The robot adjusts its wiping path on a mannequin to maintain a safe distance from an intruding hand within the workspace. 
    \textbf{ (c)} Cluttered Cleaning: The robot successfully navigates around multiple static obstacles during a whiteboard cleaning task.
    }
    \label{fig:franka_HRI}
\end{figure*}

\subsection{Results and Analysis}
\label{subsec:results}

\subsubsection{Robustness to Perturbations}
\label{subsubsec:perturbations}
Figure~\ref{fig:perturb} illustrates the response of each method to external perturbations of varying magnitudes. The NODE--CLF--CBF  method (Fig.~\ref{fig:perturb}(a)) reacts strongly
to small perturbations, accelerating away from the nominal trajectory and taking the longest time to settle; under larger perturbations, it deviates further, overshoots the goal, and produces large trajectory errors. The DMP--APF method (Fig.~\ref{fig:perturb}(b)) exhibits the sharpest deviation along perturbation, but recovers faster than NODE--CLF--CBF . In contrast, the proposed SafeDMPs method (Fig.~\ref{fig:perturb}(c)) demonstrates smooth and bounded recovery, maintaining its trajectory within a spatio-temporal tube and converging back to the nominal path most quickly.

\subsubsection{Safety from Static and Dynamic Obstacles}
\label{subsubsec:obstacle_avoidance}
The obstacle avoidance capabilities are shown in Figures~\ref{fig:obstacle_comparison} (static) and~\ref{fig:moving_obstacle_comparison} (dynamic). NODE--CLF--CBF (Figs.~\ref{fig:obstacle_comparison}(a),~\ref{fig:moving_obstacle_comparison}(a)) successfully avoids collisions but often with minimal clearance, leaving little tolerance for execution errors or modeling uncertainties. Increasing the clearance parameter improves margins but results in abrupt, jerky motion, reflecting a trade-off between smoothness and safety. DMP--APF (Figs.~\ref{fig:obstacle_comparison}(a),~\ref{fig:moving_obstacle_comparison}(b)) suffers from local minima, frequently resulting in collisions or erratic, oscillatory paths. SafeDMPs (Figs.~\ref{fig:obstacle_comparison}(c),~\ref{fig:moving_obstacle_comparison}(c)) consistently generates smooth, safe detours around both static and dynamic obstacles while maintaining a well-defined safety margin and stable convergence to the goal.

A summary of the comprehensive qualitative and quantitative comparison is presented in Table~\ref{tab:results}. The proposed SafeDMPs method demonstrates superior performance across all metrics, successfully combining formal safety guarantees with robust, efficient, and stable motion generation. Notably, the obstacle avoidance convergence time for the DMP+APF baseline is omitted due to its high collision rate, which prevented a meaningful measurement.

\subsection{Hardware Validation Results}
\label{subsec:hardware_results}

To validate SafeDMP's real-world applicability, we deployed it on a Franka Emika Panda robot operating at 1~kHz. A mannequin emulated human–robot interaction scenarios (Fig.~\ref{fig:franka}). Task training data, collected via kinesthetic demonstrations, was temporally resampled and low-pass filtered prior to DMP learning. For real-time intervention detection, an Intel RealSense camera combined with Mediapipe pose detection tracked the mannequin's hand. To account for vision-based sensing latency and tracking noise, conservative STT safety margins were incorporated to maintain tube invariance against bounded estimation errors. Ultimately, the high-frequency control loop and closed-form modulation rapidly compensate for perception jitter, ensuring collision-free behavior under realistic uncertainties.

\subsubsection{Demonstrated Tasks and Results}
\label{subsubsec:hri_tasks}

We evaluated the proposed framework through three distinct human-robot interaction (HRI) scenarios, each involving intentional human interventions to test safety and reactivity:

\begin{itemize}
    \item \textbf{Pick-and-Place with Dynamic Obstruction:} The robot executed a standard pick-and-place operation while a moving human hand randomly continued to obstruct its nominal path. SafeDMPs generated smooth trajectory deviations to reactively avoid unsafe contact while ensuring successful task completion (Fig.~\ref{fig:franka_HRI}(a)).

    \item \textbf{Assistive Wiping with Intermittent Obstruction:} The robot performed a wiping motion on a surface while a health worker randomly blocked its path intermittently. The framework reliably deformed the trajectory to maintain a safe clearance and seamlessly resumed contact-based wiping once the obstruction was removed (Fig.~\ref{fig:franka_HRI}(b)).

    \item \textbf{Whiteboard Cleaning with Cluttered Obstacles:} The robot performed a whiteboard cleaning task in the presence of multiple static obstacles intentionally placed close to its nominal path. SafeDMPs adapted the end-effector's trajectory to maneuver around the obstructions while respecting predefined safety margins and maintaining task progress (Fig.~\ref{fig:franka_HRI}(c)).
\end{itemize}
Across all hardware demonstrations, SafeDMPs exhibited the same safe interactions and reliably converged back to its nominal path, confirming the framework’s robustness in real-world, human-interactive environments (Figs.~\ref{fig:franka_HRI}(a)--\ref{fig:franka_HRI}(c)).

\section{Conclusion}
\label{sec:conclusion}

Our modular framework, \textbf{SafeDMPs}, generates safe and adaptive motion by integrating Dynamic Movement Primitives with Spatio-Temporal Tubes. This approach avoids online constrained optimization and leverages closed-form safety guarantees. SafeDMPs achieves real-time execution at high control frequencies while ensuring provable collision avoidance.

A comprehensive evaluation in simulation and on a physical Franka Emika Panda robot demonstrated that SafeDMPs consistently outperforms both optimization-based and heuristic baselines. Compared to NODE--CLF--CBF, SafeDMPs reduced execution time by up to \textbf{99.97\%}, lowered memory footprint by \textbf{65\%}, and decreased nominal trajectory error by \textbf{95\%} and perturbation error by \textbf{86\%}. Furthermore, SafeDMPs achieved up to a \textbf{94\%} faster convergence under perturbations and \textbf{35\%} faster convergence during obstacle avoidance. Unlike DMP--APF, SafeDMPs avoided oscillations and local minima, ensuring smooth and safe trajectories even in cluttered or dynamic environments.

The combination of formal safety guarantees, computational efficiency, and robustness makes SafeDMPs well-suited for deployment in collaborative, assistive, and industrial robotics, where reliable real-time adaptation in human-centered environments is essential.
\vspace{-0.3em}
\section{Limitations and Future Work}

While the proposed framework demonstrates strong performance in real-time, adaptive, and safe motion generation, several limitations remain. The current implementation relies on manual parameter tuning, which may limit scalability and generalization. Additionally, the control behavior near the boundaries of the STT produces increasingly strong corrective actions as the state approaches the tube limits, theoretically tending toward infinite control to enforce invariance, which cannot be realized due to actuator constraints.

Future work will investigate adaptive parameter selection strategies that automatically adjust tube parameters and control gains based on task context, perception uncertainty, and observed disturbances. We also plan to develop bounded-control formulations and tube redirection strategies using a world-model-informed planner to generate new feasible tubes when large disturbances occur. Furthermore, we aim to extend the framework to incorporate whole-body safety and integrate high-level spatio-temporal logic planners for handling complex task specifications and bi-arm coordination. 
\vspace{-1em}
\bibliographystyle{IEEEtran}
\bibliography{references}

\end{document}

%% file: intro.tex
For robots to seamlessly integrate into human-centric environments, they must operate with a high degree of robustness and safety. Robustness is the capacity to maintain stable, goal-directed behavior despite external disturbances, while safety is the guarantee of avoiding harm to people and property. While these two properties are paramount for real-world deployment, achieving them simultaneously in a computationally efficient manner remains a central challenge in robotic control.
\begin{figure}[htpb]
    \centering
    \includegraphics[width=0.85\linewidth]{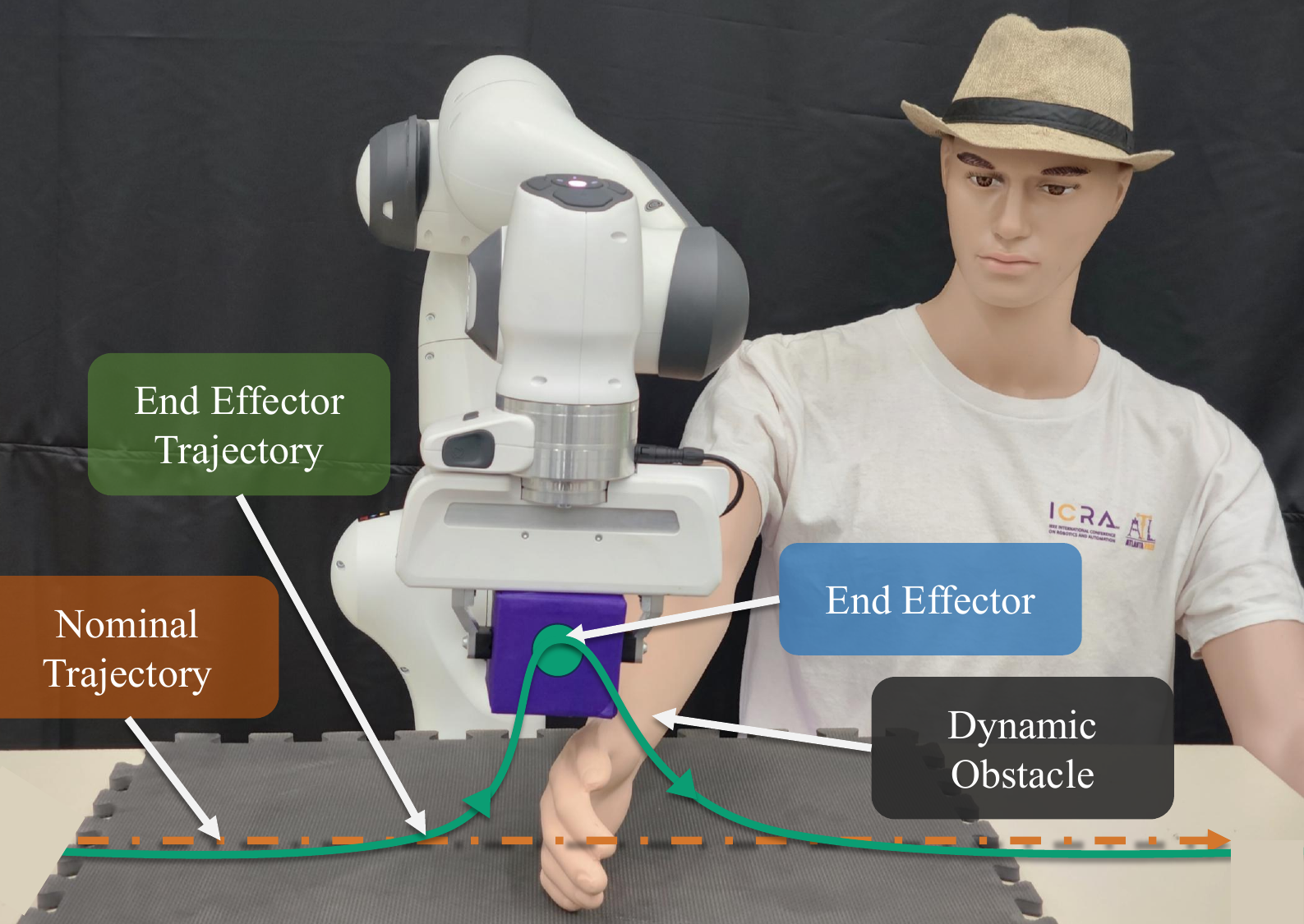}
    \caption{Hardware setup featuring the Franka  Research 3 robot equipped for human-robot interaction tasks. Illustration of SafeDMPs for adaptive Human Robot Interaction (HRI) in response to dynamic interference.}
    \label{fig:franka}
\end{figure}

One major line of research addresses robustness and generalization through Learning from Demonstration (LfD) using movement primitives. Among these, Dynamic Movement Primitives (DMPs) \cite{dmp} are particularly effective. DMPs encode trajectories as attractor-based dynamical systems, which provides inherent stability and robustness to perturbations from a single demonstration. This offers a distinct advantage over methods like Gaussian Mixture Models (GMMs) \cite{gmm} or Probabilistic Movement Primitives (ProMPs) \cite{prodmp2}, which often require numerous demonstrations and can be computationally complex. Other approaches, such as Kernelized Movement Primitives (KMPs) \cite{KernelizedMovementPrimitives} or task-space deformation techniques \cite{li2023task}, can adapt motions but require explicit, user-defined constraints that are often difficult to specify. Despite their strengths in robust trajectory encoding, DMPs fundamentally lack a principled mechanism for safety. They are typically paired with reactive, heuristic methods like Artificial Potential Fields (APFs) \cite{dmp_based_oa}, which are notoriously susceptible to local minima and unstable oscillations, providing no formal safety guarantees.

With stability in mind, research in this field has evolved to provide provable guarantees using formal methods. For stability, while DMPs provide it inherently, other sophisticated learning frameworks like Stable-BC \cite{mehta2025stable}, Lyapunov Density Models \cite{kang2022lyapunov}, and Euclideanizing Flows \cite{rana2020euclideanizing, perez2023stable} have been developed. However, these methods typically require extensive training over multiple demonstrations. For safety, the use of Control Barrier Functions (CBFs) has become a cornerstone for guaranteeing forward invariance of a safe set. Frameworks combining Dynamical Systems \cite{khansari2011learning, figueroa2018physically} with Control Lyapunov Functions (CLFs) and CBFs, such as NODE-CLF-CBF \cite{nawaz2024learning}, can enforce provable safety and stability. The critical drawback of this class of methods is their reliance on solving a constrained quadratic program at every control step. This online optimization imposes a significant computational burden, leading to sluggish, non-smooth behavior unsuitable for high-frequency control. While alternative closed-form safety methods like the Rotational Obstacle Avoidance Method (ROAM) \cite{huber2023avoidance} exist, they offer less explicit control over the safety margin, leading to less predictable path shapes compared to CBFs.

The literature presents a clear trade-off: efficient and robust trajectory generation methods, like DMPs, that lack formal safety guarantees, versus formally safe methods that are computationally expensive due to online optimization. This dichotomy highlights a core challenge in robotics, i.e., balancing performance with provable safety. This paper presents SafeDMP, a novel framework that resolves this trade-off by marrying the robustness and efficiency of DMPs with the provable safety of formal methods, without the computational overhead of online optimization.

Our core contribution is the integration of Spatio-Temporal Tubes (STTs), which define a time-varying safe envelope around the nominal DMP trajectory. The STT provides a closed-form, non-optimization-based control law that provably ensures the robot’s motion remains within a collision-free region. As seen in Figure~\ref{fig:franka}, the robot deviates from its nominal trajectory in the presence of human interference while maintaining a minimum safety margin, and subsequently converges back to the nominal path once the interference is removed. This synergy results in a system that is not only inherently robust to perturbations but also guaranteed to adapt safely in the presence of static and dynamic obstacles. Extensive experiments demonstrate that SafeDMP achieves orders-of-magnitude faster execution times and more accurate trajectory reproduction than optimization-based baselines, making it an ideal solution for real-time deployment in dynamic, human-centric environments.